\documentclass[11pt,a4paper]{article}
\usepackage[a4paper]{geometry}
\usepackage[nohyperref]{emnlp2017}
\usepackage[breaklinks=true]{hyperref} 
\usepackage{times}
\usepackage{latexsym}
\usepackage{graphicx}
\usepackage{amsmath}
\usepackage{multirow}
\usepackage{booktabs}
\usepackage{natbib}
\usepackage{framed}
\usepackage{pbox}

\def\emnlppaperid{***}

\let\oldhref\href
\renewcommand{\href}[2]{\oldhref{#1}{\hbox{#2}}}

\emnlpfinalcopy

\newcommand{\fulllabel}[3]{\textsc{#1}\newline\textbf{#2}\newline\textsc{#3}}

\title{The RepEval 2017 Shared Task:\\ Multi-Genre Natural Language Inference with Sentence Representations} 

\author{
Nikita Nangia$^{1}$ \\
\texttt{\small nikitanangia@nyu.edu} \\
\And
Adina Williams$^{2}$ \\
\texttt{\small adinawilliams@nyu.edu} \\
\And
Angeliki Lazaridou$^{3}$ \\
\texttt{\small angeliki@deepmind.com} \\
\And
Samuel R.~Bowman$^{1,2}$ \\
\texttt{\small bowman@nyu.edu} \\
\AND
$^{1}$\normalfont Center for Data Science\\ New York University\And
$^{2}$\normalfont Department of Linguistics\\ New York University\And
$^{3}$\normalfont DeepMind
}

\date{}

\begin{document}
\maketitle

\begin{abstract}
This paper presents the results of the RepEval 2017 Shared Task, which evaluated neural network sentence representation learning models on the Multi-Genre Natural Language Inference corpus (MultiNLI) recently introduced by \citet{williams2017broad}. All of the five participating teams beat the bidirectional LSTM (BiLSTM) and continuous bag of words baselines reported in \citeauthor{williams2017broad}. The best single model used stacked BiLSTMs with residual connections to extract sentence features and reached 74.5\% accuracy on the genre-matched test set. Surprisingly, the results of the competition were fairly consistent across the genre-matched and genre-mismatched test sets, and across subsets of the test data representing a variety of linguistic phenomena, suggesting that all of the submitted systems learned reasonably domain-independent representations for sentence meaning.

\end{abstract}

\begin{table*}[ht]
  \centering\small
  \begin{tabular}{p{6.5cm}p{2.2cm}p{6cm}}
  \toprule
Met my first girlfriend that way. & \fulllabel{Face-to-Face}{contradiction}{} & I didn't meet my first girlfriend until later.\\
\rule{0pt}{4ex}He turned and saw Jon sleeping in his half-tent. & \fulllabel{Fiction}{entailment}{} & He saw Jon was asleep.  \\
\rule{0pt}{4ex}8 million in relief in the form of emergency housing. & \fulllabel{Government}{neutral}{} & The 8 million dollars for emergency housing was still not enough to solve the problem.\\
\rule{0pt}{4ex}Now, as children tend their gardens, they have a new appreciation of their relationship to the land, their cultural heritage, and their community. & \fulllabel{Letters}{neutral}{} & All of the children love working in their gardens.\\
\rule{0pt}{4ex}At 8:34, the Boston Center controller received a third transmission from American 11 & \fulllabel{9/11}{entailment}{} & The Boston Center controller got a third transmission from American 11. \\
\rule{0pt}{4ex}In contrast, suppliers that have continued to innovate and expand their use of the four practices, as well as other activities described in previous chapters, keep outperforming the industry as a whole. & \fulllabel{OUP}{contradiction}{} & The suppliers that continued to innovate in their use of the four practices consistently underperformed in the industry.\\
\rule{0pt}{4ex}I am a lacto-vegetarian. & \fulllabel{Slate}{neutral}{} & I enjoy eating cheese too much to abstain from dairy.\\
\rule{0pt}{4ex}someone else noticed it and i said well i guess that's true and it was somewhat melodious in other words it wasn't just you know it was really funny & \fulllabel{Telephone}{contradiction}{} & No one noticed and it wasn't funny at all. \\
\rule{0pt}{4ex}For more than 26 centuries it has witnessed countless declines, falls, and rebirths, and today continues to resist the assaults of brutal modernity in its time-locked, color-rich historical center. & \fulllabel{Travel}{entailment}{} & It has been around for more than 26 centuries. \\
\rule{0pt}{4ex}If you need this book, it is probably too late' unless you are about to take an SAT or GRE. & \fulllabel{Verbatim}{contradiction}{} & It's never too late, unless you're about to take a test. \\
    \bottomrule
  \end{tabular}
  \caption{Randomly chosen examples from each genre of the MultiNLI development set.}
  \label{examples}
\end{table*}

\section{Introduction}

The Second Workshop on Evaluating Vector Space Representations for NLP (RepEval 2017) features a shared task competition meant to evaluate natural language understanding models based on sentence encoders---that is, models that transform sentences into fixed-length vector representations and reason using those representations. Submitted systems are evaluated on the task of natural language inference (NLI, also known as recognizing textual entailment, or RTE) on the Multi-Genre NLI corpus (MultiNLI; \citealt{williams2017broad}). Each example in the corpus consists of a pair of sentences, and systems must predict whether the relationship between the two sentences is \textit{entailment}, \textit{neutral} or \textit{contradiction} in a balanced three-way classification setting. 

We selected the task of NLI with the intent to evaluate as directly as possible the degree to which each model can extract and manipulate distributed representations of sentence meaning. In order for a system to perform well at natural language inference, it needs to handle nearly the full complexity of natural language understanding,\footnote{Entailment notably does not require a system to ground its representations of sentence meaning to any outside representational system, for better or worse. For related discussion of entailment and natural language understanding see \citet{chierchia::meaning::1991}, \citet{dagan2006pascal}, and \citet{MacCartney09}.} but its framing as a sentence-pair classification problem makes it suitable as an evaluation task for a broad range of models, and avoids issues of sequence generation, structured prediction, or memory access that can complicate evaluation in other settings.

The shared task includes two evaluations, a standard in-domain (\textit{matched}) evaluation in which the training and test data are drawn from the same sources, and a cross-domain (\textit{mismatched}) evaluation in which the training and test data differ substantially. This cross-domain evaluation tests the ability of submitted systems to learn representations of sentence meaning that capture broadly useful features.

This paper briefly introduces the task and dataset, presents the rules and results of the competition, and analyzes and compares the submitted systems. All the submitted systems are broadly similar, and incorporate bidirectional recurrent neural networks as a key component. We find that all systems performed fairly well, outperforming a simple bidirectional LSTM \citep[BiLSTM;][]{hochreiter1997long} baseline. To our surprise, no system performed dramatically worse on the \textit{mismatched} evaluation than on the \textit{matched} evaluation, and all systems performed reasonably consistently across examples representing a range of linguistic phenomena,  suggesting that all were able to produce systems for semantic representation which, while not perfect, were effective and not tightly adapted to any particular style of language or set of constructions.

\section{Dataset}

MultiNLI \citep{williams2017broad} consists of 393k pairs of sentences from a broad range of genres of written and spoken English, balanced across three labels. Each premise sentence (the first sentence in each pair) is derived from one of ten sources of text, which constitute the ten genre sections of the corpus. Each hypothesis sentence and pair label was composed by a crowd worker in response to a premise. MultiNLI was designed and collected in the style of the Stanford NLI Corpus (SNLI; \citealt{snli:emnlp2015}), but covers a broader range of styles of text, rather than the relatively homogeneous captions used in SNLI.

Testing and development sets are available for all genres, with 2000 examples per set per genre. Only five genres have accompanying training sets. So, for the \textit{matched} development and test sets, models are tested on examples derived from the same sources as those in the training set, while for the \textit{mismatched} sets, the text source is not represented in the training data. 

Table~\ref{examples} presents example sentences from the corpus and Table~\ref{stats} presents some key statistics. For a detailed discussion of the corpus, refer to \citet{williams2017broad}.

\begin{table*}[t]
  \centering\small 
  \begin{tabular}{lrrrcrrrcr}
  \toprule
\bf  & \multicolumn{3}{c}{\bf \#Examples} & \bf \#Wds. & \multicolumn{2}{c}{\bf `S' parses} &\\ 
\bf Genre & \bf Train& \bf Dev. & \bf Test & \bf Prem. & \bf Prem. & \bf Hyp. &  \bf Agrmt.& \bf BiLSTM Acc. \\
\midrule
\it SNLI & \it 550,152 & \it 10,000 & \it 10,000 & \it 14.1 & \it 74\%& \it 88\% & \it 89.0\% & \it 81.5\% \\
\midrule
\sc Fiction & 77,348 & 2,000 & 2,000 & 14.4 & 94\%&97\% & 89.4\%& 66.8\%  \\
\sc Government & 77,350 & 2,000 & 2,000 & 24.4 &90\%&97\%& 87.4\%& 68.0\%  \\
\sc Slate & 77,306 & 2,000 & 2,000 & 21.4 &94\%&98\%& 87.1\%& 68.4\% \\
\sc Telephone & 83,348 & 2,000 & 2,000 & 25.9 &71\%&97\%& 88.3\%& 67.7\% \\
\sc Travel & 77,350 & 2,000 & 2,000 & 24.9 &97\%&98\%& 89.9\%& 66.8\% \\
\midrule
\sc 9/11 & 0 & 2,000 & 2,000 & 20.6 &98\%&99\%& 90.1\%& 68.5\% \\
\sc Face-to-face & 0 & 2,000 & 2,000 & 18.1 &91\%&96\%& 89.5\%& 67.5\% \\
\sc Letters & 0 & 2,000 & 2,000 & 20.0 &95\%&98\%& 90.1\%& 66.4\% \\
\sc OUP & 0 & 2,000 & 2,000 & 25.7 &96\%&98\%& 88.1\%& 66.7\%  \\
\sc Verbatim & 0 & 2,000 & 2,000 & 28.3 &93\%&97\%& 87.3\%& 67.2\% \\
\midrule
\bf MultiNLI Overall & \textbf{392,702} & \textbf{20,000} & \textbf{20,000} & \bf 22.3 & \bf 91\%&\bf 98\% & \bf 88.7\% & \bf 67.4\% \\
    \bottomrule
  \end{tabular}
  \caption{ Key statistics for the corpus broken down by genre, presented alongside figures from SNLI for comparison. The first five genres represent the \textit{matched} section of the development and test sets, and the remaining five represent the \textit{mismatched} section. The first three statistics shown are the number of examples in each genre. \textit{\#Wds. Prem.}~is the mean token count among premise sentences. \textit{`S' parses} is the percentage of premises or hypotheses which the Stanford Parser labeled as full sentences rather than fragments. \textit{Agrmt.}~is the percent of individual annotator labels that match the assigned gold label used in evaluation. \textit{BiLSTM Acc.}~gives the test accuracy on the full test set for the BiLSTM baseline model trained on MultiNLI and SNLI.}
  \label{stats}
\end{table*}

\section{Shared Task Competition}

The purpose of the shard task is to evaluate techniques for training and using sentence encoders. To this end, we require that all models create fixed-length vectors for each sentence with no explicitly-imposed internal structure. Alignment strategies like attention that pass information between the two encoders handling the two input sentences in a pair are not allowed. Memory models that represent sentences as variable-length sets or sequences of vectors are also not permitted. While systems that use methods like attention and structured memory are effective for NLI \citep[][i.a.]{rocktaschel2015reasoning,wang2015learning,chen2017esim,williams2017broad}, much of the variation across models of this kind lies in the way that they explicitly or implicitly align related sentences, rather than the way that they extract representations for sentences. As a result, we expect that focusing our evaluation on a restricted subset of models will yield conclusions that are more generally applicable to work on natural language understanding than would have been the case otherwise.

\paragraph{Additional Rules} We provide competitors with labeled training and development sets, and unlabeled test sets for which they must submit labels. The development sets are meant to be used for hyperparameter tuning and model selection, and training on the development sets is not allowed. We place no limits on the use of outside training data and resources except that they be publicly available. We specifically encourage the use of the SNLI training set. Multiple submissions from the same team are allowed, up to a limit of two per day during the two-week evaluation period. Individual participants (i.e., PIs) are permitted to join multiple teams within reason, but only when each team reflects a fully independent engineering effort and each team has a different lead developer. 

\paragraph{Evaluation} Competitors had approximately ten weeks, starting with the release of the MultiNLI training and development sets, to develop their systems and two additional weeks---the evaluation period---to run their systems on the unlabeled test sets and submit results. The shared task evaluation was hosted through the Kaggle in Class platform using two competition pages---one each for the \textit{matched}\footnote{\url{https://inclass.kaggle.com/c/multinli-matched-evaluation}} and \textit{mismatched}\footnote{\url{https://inclass.kaggle.com/c/multinli-mismatched-evaluation}} sections of the corpus. The public leaderboard, which was displayed during the evaluation period, showed results on a random 25\% of the test set labels, and the final results were computed by evaluating the two best systems from each competitor (chosen from the public leaderboard) on the remaining hidden 75\% of the test set labels. 

\begin{table*}[t]\small
  \centering\small 
  \begin{tabular}{llrrl}
  \toprule
  
    \bf Team Name & \bf Authors & \bf Matched & \bf Mismatched & \bf Model Details \\
    \midrule
    alpha (ensemble) &  \citeauthor{chen2017gated}~&  \bf 74.9\% & \bf 74.9\% & \textsc{Stack, Char, Attn., Pool, ProdDiff} \\
    YixinNie-UNC-NLP &  \citeauthor{nie2017stacked}~& \underline{74.5\%}  & \underline{73.5\%} & \textsc{Stack, Pool, ProdDiff, SNLI}\\
    alpha & \citeauthor{chen2017gated}~&  73.5\%  & \underline{73.6\%} & \textsc{Stack, Char, Attn, Pool, ProdDiff} \\
    Rivercorners (ensemble) & \citeauthor{balazs2017refining}~& 72.2\% & 72.8\% & \textsc{Attn, Pool, ProdDiff, SNLI}\\
    Rivercorners &  \citeauthor{balazs2017refining}~& 72.1\%  & 72.1\% & \textsc{Attn, Pool, ProdDiff,  SNLI} \\
    LCT-MALTA & \citeauthor{vu2017LCT}   & 70.7\% & 70.8\% & \textsc{Char, EnhEmb, ProdDiff, Pool}\\
    TALP-UPC &  \citeauthor{yang2017cian} & 67.9\%  & 68.2\% & \textsc{Char, Attn, SNLI} \\
    BiLSTM baseline & Williams et al. & 67.0\%  & 67.6\% & \textsc{Pool, ProdDiff, SNLI}\\
    \bottomrule
    
   \end{tabular}
   \caption{RepEval 2017 shared task competition results. The Model Details column lists some of the key strategies used in each system, using keywords: \textsc{Stack:} use of multilayer bidirectional RNNs, \textsc{Char:} character-level embeddings, \textsc{EnhEmb:} embeddings enhanced with auxiliary features, \textsc{Pool:} max or mean pooling over RNN states, \textsc{Attn:} intra-sentence attention, \textsc{ProdDiff:} elementwise sentence product and difference features in the final entailment classifier, \textsc{SNLI:} use of the SNLI training set.} 
   \label{results} 
\end{table*}

\section{Results and Leaderboard}
The competition results are shown in Table~\ref{results}. All evaluated systems beat the BiLSTM baseline reported in \citeauthor{williams2017broad}. Furthermore, there is only a marginal gap between accuracy on \textit{matched} and \textit{mismatched} test sets for all systems.

The best performing single model is by \citeauthor{nie2017stacked}, who achieve the best result on the \textit{matched} competition and tie with \citeauthor{chen2017gated}~in the \textit{mismatched} competition. The \citeauthor{nie2017stacked}~model architecture uses stacked BiLSTMs with residual connections and, unlike the other high performing models, does not use within-sentence attention. The best performing system overall is an ensemble by \citeauthor{chen2017gated}, which is based closely on the Enhanced Sequential Inference Model \citep[ESIM;][]{chen2017esim} but with attention only within each sentence, rather than between the two.


Looking toward the future, we also made available non-time-limited Kaggle in Class competition pages\footnote{Matched: \url{https://inclass.kaggle.com/c/multinli-matched-open-evaluation}\\ Mismatched: \url{https://inclass.kaggle.com/c/multinli-mismatched-open-evaluation}} to allow for further fair evaluations on the MultiNLI test sets. Note that since these evaluation sites report results on 100\% of the test set, rather than the 75\% used in the shared task, numbers reported on that site may differ slightly from those seen in the competition.

\section{Model Comparison}
All of the submitted systems are based on bidirectional LSTMs, but each system uses this core tool in a somewhat different way. This section surveys the key differences between systems, and the Model Details column in Table~\ref{results} serves as a summary reference for these differences. 

\paragraph{Depth} \citeauthor{chen2017gated}~and \citeauthor{nie2017stacked} use three-layer bidirectional RNNs, while others only used single-layer RNNs. This likely contributes significantly to their good performance, as it is the most prominent feature shared only by these two top systems. They both use shortcut connections between recurrent layers to ease gradient flow, and \citeauthor{nie2017stacked}~find in an ablation study that using shortcut connections improves their performance by over 1\% on both development sets.

\paragraph{Embeddings} Systems vary reasonably widely in their approach to input encoding. 
\citeauthor{yang2017cian}~and \citeauthor{chen2017gated}~use a combination of GloVe embeddings \citep[][not fine tuned]{pennington2014glove} and character-level convolutional neural networks \citep{kim2015charCNN} to extract representations of words. 
\citeauthor{balazs2017refining}~also use pre-trained GloVe embeddings without fine tuning, but report (contra \citeauthor{chen2017esim}) that an added character-level feature extractor does not improve performance.

\citeauthor{vu2017LCT}~use pre-trained GloVe word embeddings augmented with additional feature vectors. They create embeddings for part-of-speech (POS), character level information, and the dependency relation between a word and its parent, and concatenate these with the embedding for each word. They find that this supplies a small but nontrivial improvement to their development set performance, especially in the \textit{mismatched} setting.

\citeauthor{nie2017stacked}~use the simplest strategy, initializing embeddings with GloVe vectors and fine-tuning them.

\paragraph{Pooling}
\citeauthor{vu2017LCT}~make a surprisingly effective change to the baseline BiLSTM model, motivated by \citeauthor{conneau:2017:supervised}'s \citeyearpar{conneau:2017:supervised} findings, by using max pooling rather than mean pooling when collecting the hidden states of the bidirectional LSTM for use as a sentence representation. They find that this yields an improvement of over 2.5\% on both development sets.

While \citeauthor{vu2017LCT} show that the choice of pooling strategy is quite important, \citeauthor{balazs2017refining}~do not find a substantial effect in a similar comparison. This may be because \citeauthor{balazs2017refining}'s model also makes use of intra-sentence attention following the pooling layer, which dramatically reduces the importance of pooling.

\paragraph{Intra-Sentence Attention}
\citeauthor{chen2017gated}~and \citeauthor{balazs2017refining}~both use attention over the BiLSTM states of each sentence to compute a final representation for that sentence. \citeauthor{chen2017gated}~in particular uses a novel \textit{gated} attention formulation, in which the BiLSTM gate values supply the attention weights over hidden states according to
$$ v_g = \sum_{i=1}^n \frac{\lVert g_i \rVert_2}{\sum_{j=1}^n \lVert g_j \rVert_2} h_i$$
where $g_i$ is the BiLSTM input gate and $h_i$ is the output from the BiLSTM encoder. They find that their use of gated attention helps performance somewhat relative to an unspecified baseline, though only in the \textit{matched} setting.
 
\paragraph{Sentence Pair Classifier} 
Every system but \citeauthor{yang2017cian}'s uses elementwise product and difference features, comparing the two sentence encodings as part of the input to the classifier MLP that predicts the final relation label. In an ablation study, \citeauthor{chen2017gated} find this to be highly important, yielding more than a 3\% gain in performance on both development sets.

\paragraph{Data and SNLI}

We observe relatively little variation in the training data used in submitted systems. All systems are trained only on labeled NLI data---either the MultiNLI training set alone, or the MultiNLI and SNLI training sets combined. While \citeauthor{williams2017broad} find that the combined training set yields somewhat better results on the MultiNLI test set, \citeauthor{chen2017esim} nonetheless reaches state-of-the-art performance without using it.

\paragraph{Interim Discussion}

We were particularly struck by the effectiveness of the max pooling strategy as a simple and highly effective improvement to the baseline BiLSTM sentence encoder. Less surprisingly, depth and intra-sentence attention appear to be broadly effective, and product and difference features appear to be valuable when using sentence encoders for the task of NLI. The results surrounding embeddings and input encoding were less clear, though \citeauthor{nie2017stacked}'s use of pre-trained GloVe embeddings with fine tuning appears to be a simple and effective approach.

\section{Error Analysis}
In the interest of better understanding both the corpus and the submitted models, we annotate a 1,000-sample subset of the development set. We also provide a set of probe sentences and ask participating teams to submit vectors for all sentences in the probe set and test set. This section surveys our methods findings.

\subsection{Annotations}

\begin{table*}[t]
  \centering\small
  \begin{tabular}{llrrrrr}
  \toprule
    & \bf Annotation Tag & \bf Label Frequency & \bf BiLSTM & \bf Yang & \bf Balazs (S) & \bf Chen (S)\\
   \midrule
   \multirow{13}{*}{Matched}
   & CONDITIONAL       & 5\% & 100\% & 100\% & 100\% & 100\% \\ 
   & WORD\_OVERLAP        & 6\% & 50\% & 63\% & 63\%  & 63\% \\ 
   & NEGATION           & 26\% & 71\% & 75\% & 75\%  & 75\%  \\ 
   & ANTO       & 3\% & 67\% & 50\% & 50\%  & 50\%  \\ 
   & LONG\_SENTENCE      & 20\% & 50\% & \bf 75\% & \bf 75\%  & 67\%  \\ 
   & TENSE\_DIFFERENCE    & 10\% & 64\% & 68\% & 68\%  & \bf 86\% \\ 
   & ACTIVE/PASSIVE      & 3\% & 75\% & 75\% & 75\%  & 88\% \\ 
   & PARAPHRASE      & 5\% & 78\% & 83\% & 83\%  & 78\% \\
   & QUANTITY/TIME\_REASONING  & 3\% & 50\% & 50\% & 50\% & 33\% \\ 
   & COREF  & 6\% & 83\% & 83\% & 83\%  & 83\%  \\ 
   & QUANTIFIER & 25\% & 64\% & 59\% & 59\%  & \bf 74\%   \\
   & MODAL  & 29\% & 66\% & 65\% & 65\%  & \bf 75\%   \\ 
   & BELIEF  & 13\% & 74\% & 71\% & 71\%  & 73\%  \\ \midrule
    
   \multirow{13}{*}{Mismatched}
    & CONDITIONAL & 5\% & 100\% & 80\% & 80\% & 100\% \\
    & WORD\_OVERLAP & 7\% & 58\% & 62\% & 62\% & 76\%   \\
    & NEGATION & 21\%  & 69\% & 73\% & 73\%  & 72\%   \\
    & ANTO & 4\% & 58\% & 58\% & 58\%  & 58\%   \\
    & LONG\_SENTENCE & 20\% & 55\% &  67\% &  67\%  & 67\%  \\
    & TENSE\_DIFFERENCE & 4\% & 71\% & 71\% & 71\%  & 89\%   \\
    & ACTIVE/PASSIVE & 2\% & 82\% & 82\% & 82\%  & 91\%   \\
    & PARAPHRASE & 7\% & 81\% & 89\% & 89\%  & 89\%   \\
    & QUANTITY/TIME\_REASONING & 8\% & 46\% & 54\% & 54\%  & 46\%  \\
    & COREF & 6\% & 80\% & 70\% & 70\%  & 80\% \\
    & QUANTIFIER  & 28\% & 70\% & 68\% & 68\%  & \bf 77\%   \\
    & MODAL & 25\% & 67\% & 67\% & 67\%  & \bf 76\%   \\
    & BELIEF & 12\% & 73\% & 71\% & 71\%  & 74\%   \\ \bottomrule

   \end{tabular}
   \caption{This table shows the accuracy of different models for each tagged subset of our 1,000-example development set sample. The `(S)' indicates that results for the single model are shown. Some results that stand out to us are shown in bold.}
   \label{tags} 
\end{table*}

The annotated subset of the development set was released to competitors during the model development period, and consists of 1,000 examples each tagged with zero or more of the following labels. Labels were assigned manually except where clear keyword-spotting techniques sufficed. 

\begin{itemize}
    \item \textsc{conditional}: Whether either sentence contains a conditional.\\
    Example: \textbf{P:} \textit{Laser-cutting equipment must be totally enclosed to be safe for human operators.} \textbf{H:} \textit{Even \underline{if} the laser machine is fully contained within, there still exist some amount of risk for the workers in the close proximity.}

	\item \textsc{active/passive}: Whether there is an active-to-passive (or vice versa) transformation from the premise to the hypothesis.\\ 
	Example: \textbf{P:} \textit{Hani Hanjour, Khalid Al Mihdhar, and Majed Moqed \underline{were flagged by} capps.} \textbf{H:} \textit{Capps never \underline{flagged} anyone.}

	\item \textsc{paraphrase}: Whether the two sentences are close paraphrases.\\ 	Example: \textbf{P:} \textit{Uh, lets see.} \textbf{H:} \textit{Let us look.}

	\item \textsc{coref}: Whether the hypothesis contains a pronoun or referring expression that needs to be resolved using the premise. \\
	Example: \textbf{P:} \textit{You and I, gentle reader, are accredited members of the guild.} \textbf{H:} \textit{\underline{We} are recognised as members of the guild.}

	\item \textsc{quantifier}: Whether either sentence contains one of the following quantifiers: \textit{much, enough, more, most, less, least, no, none, some, any, many, few, several, almost, nearly}.\\
	Example: \textbf{P:} \textit{We have provided an invoice to facilitate your gift.} \textbf{H:} \textit{There's \underline{no} invoice available for your gift.}

	\item \textsc{modal}: Whether either sentence contains one of the following modal verbs: 
	\textit{can, could, may, might, must, will, would, should.} \\
	Example: \textbf{P:} \textit{Conversely, an increase in government saving adds to the supply of resources available for investment and \underline{may} put downward pressure on interest rates.} \textbf{H:} \textit{The amount of resources available for investment increases when government savings are increased.}
	
	\item \textsc{belief}: Whether either sentence contains one of the following belief verbs: \textit{know, believe, understand, doubt, think, suppose, recognize, recognize, forget, remember, imagine, mean, agree, disagree, deny, promise}. \\
	Example: \textbf{P:} \textit{I trust that this is a fillip of propaganda and not a serious query.} \textbf{H:} \textit{I \underline{believe} this is to get attention and not a real inquiry.}

	\item \textsc{negation}: Whether either sentence contains negation.\\ 
	Example: \textbf{P:} \textit{On reflection, the parts will hold together.} \textbf{H:} \textit{The parts will not hold together.}

	\item \textsc{anto}: Whether the two sentences contain an antonym pair.\\
	 Example: \textbf{P:} \textit{As united 93 left Newark, the flight's crew members were \underline{unaware} of the hijacking of American 11.} \textbf{H:} \textit{As the flight United 93 left Newark the crew members were fully \underline{aware} of the hijacking of American 11.}
	 
	\item \textsc{tense\_difference}: Whether the two sentences use different tenses on any verbs.\\
	Example: \textbf{P:} \textit{Does she like what she \underline{does}?} \textbf{H:} \textit{Does she like what she \underline{is doing}?}
	
	\item \textsc{quantity/time\_reasoning}: Whether understanding the pair requires quantity or time reasoning.\\
	Example: \textbf{P:} \textit{The vice chairman joined the conference shortly before 10:00; the secretary, shortly before 10:30.} \textbf{H:} \textit{The secretary joined before the vice chairman.}

	\item \textsc{word\_overlap}: Whether the two sentences share more than 70\% of their tokens. \\
	Example: \textbf{P:} \textit{Let's look for paua shells!} \textbf{H:} \textit{Let's look for sticks.}

	\item \textsc{long\_sentence}: Whether the premise or hypothesis is longer than 30 or 16 words respectively.\\ 
	Example: \textbf{P:} \textit{As invested with its dignity, since the seventeenth century just as the crown has been used for the monarch, or the oval office has come to stand for the president of the United States.} \textbf{H:} \textit{Nobody in Britain associates the crown with the monarchy.}

\end{itemize}

Table~\ref{tags} shows model results on tagged examples for the BiLSTM baseline and for the three systems for which we were able to acquire example-by-example development set results (submission of these results was optional). Among those tags that are frequent enough to yield clearly interpretable numbers, none indicates a subset of the corpus that is dramatically harder or easier for the submitted models than is the corpus overall. This suggests that---as is typical with neural network models---these models do not rely strongly on any particular structural properties of the input texts to the exclusion of others.

We note that the submitted systems that use intra-attention (the three shown) do relatively well on the \textsc{long\_sentence} and \textsc{negation} tags. This technique likely helps the encoders to recover the structures of long sentences and to correctly identify the scope of instances of negation. We also note that all systems do relatively poorly on the \textsc{quantity/time\_reasoning} section, suggesting that these simple sentence feature extractors are not well situated to learn quantitative reasoning in this setting.

\begin{table}
  \centering
  \begin{tabular}{lr}
  \toprule
    Authors & 1-NN Genre Accuracy \\
    \midrule
    \citeauthor{chen2017gated}~& 67.3\% \\
    \citeauthor{nie2017stacked}~& 74.0\% \\ 
    \citeauthor{balazs2017refining}~& 69.2\% \\
    \citeauthor{vu2017LCT}~& 67.0\% \\
    \citeauthor{yang2017cian}~& 54.7\% \\
  \bottomrule
   \end{tabular}
   \caption{A thousand sentences are randomly sampled from the \textit{matched} test set and their pairwise distances to all sentences in the test set (premises and hypotheses) are calculated. This table shows the percentage of times the first nearest neighbor belongs to the same genre as the sample sentence.}   \label{1nn} 
\end{table}

\begin{table*}
  \centering\small
  \begin{tabular}{p{3cm}p{0.7cm}p{.4cm}p{10cm}}
  \toprule
  
    \bf Sample &\bf Model & \multicolumn{2}{l}{\bf Nearest Neighbours} \\
    \midrule
    \multirow{15}{*}{\vspace{-2.5cm} \pbox{2.9cm}{Students love the rich culture and history of the school. \textsc{(tr.)}}} 
    & \multirow{3}{*}{\vspace{.65cm}Chen} 
    & \textsc{tel.} & my son loved learning about computers in high school   \\
    \rule{0pt}{2.5ex}& & \textsc{tr.} & Families love this city-within-a-city on the beach.  \\
    \rule{0pt}{2.5ex}& & \textsc{tr.} &  The urban working class loved the new factories. \\
    
    \rule{0pt}{4ex}& \multirow{3}{*}{\vspace{.65cm}Nie} 
    &  \textsc{tel.}  & my son loved learning about computers in high school   \\
    \rule{0pt}{2.5ex}& & \textsc{tel.} &  I really loved it when I was in middle school.\\
    \rule{0pt}{2.5ex}& & \textsc{sl.} & A librarian and fellow patient kindled his love for literature more than school.  \\
    
    \rule{0pt}{4ex}& \multirow{3}{*}{\vspace{.65cm}Balazs} 
    & \textsc{sl.} & School, more than anything else, was credited for his love of literature. \\
    \rule{0pt}{2.5ex}& & \textsc{sl.} &  A librarian and fellow patient kindled his love for literature more than school. \\
    \rule{0pt}{2.5ex}& & \textsc{tel.} & I really loved it when I was in middle school. \\
    
    \rule{0pt}{4ex}& \multirow{3}{*}{\vspace{.65cm}Vu} 
    & \textsc{sl.} & A librarian and fellow patient kindled his love for literature more than school. \\
    \rule{0pt}{2.5ex}& & \textsc{tr.} & France's oldest city is a wonderful destination, with rich history and extreme beauty.  \\
    \rule{0pt}{2.5ex}& & \textsc{fic.} & The rave had some of the best artists and celebrities.   \\
    
    \rule{0pt}{4ex}& \multirow{3}{*}{\vspace{.65cm}Yang} 
    & \textsc{tr.} & The urban working class loved the new factories. \\
    \rule{0pt}{2ex}& & \textsc{fic.} & my son loved learning about computers in high school \\
    \rule{0pt}{2ex}& & \textsc{tr.} &  This area is a favorite of hikers who enjoy invigorating journeys through dense forests and along the river valleys celebrated in the paintings of Gustave Courbet.
    \smallskip\\ 
    
    \midrule
    
   \multirow{15}{*}{\vspace{-6.8cm} \pbox{2.9cm}{Critics loved Merchant-Ivory's final movie, which was an adaption of a novel written by Kaylie Jones. (\textsc{sl.})}}
    & \multirow{3}{*}{\vspace{.65cm}Chen}
    & \textsc{sl.} &  Critics laud Merchant-Ivory's exit from the 19th century in this adaptation of a semiautobiographical novel by Kaylie Jones (daughter of novelist James Jones). \\
    \rule{0pt}{2.5ex}& & \textsc{sl.} & I loved Begnigni's movie! \\
    \rule{0pt}{2.5ex}& &  \textsc{sl.} & Mercer was the lifelong love of Franklin Roosevelt, and the revelation of their affair nearly ended his marriage to Eleanor. \\
    
    \rule{0pt}{4ex}& \multirow{3}{*}{\vspace{.65cm}Nie} 
    & \textsc{sl.} & Critics laud Merchant-Ivory's exit from the 19th century in this adaptation of a semiautobiographical novel by Kaylie Jones (daughter of novelist James Jones). \\
    \rule{0pt}{2.5ex}& &  \textsc{sl.} & Critics find the book entertaining, praising digressions on gambling, laughing, and love, as well as Pinker's pop-culture references.\\
    \rule{0pt}{2.5ex}& & \textsc{sl.} & Critics think that Lichtenstein was a contemporary genius. \\
    
    \rule{0pt}{4ex}& \multirow{3}{*}{\vspace{.65cm}Balazs} 
    &  \textsc{sl.} & Critics laud Merchant-Ivory's exit from the 19th century in this adaptation of a semiautobiographical novel by Kaylie Jones (daughter of novelist James Jones). \\
    \rule{0pt}{2.5ex}& &  \textsc{tel.} & The period of the civil war is very interesting to me, I've read about 3 novels about that, including John Jakes ones. \\
    \rule{0pt}{2.5ex}& & \textsc{sl.} & The most vivid moments in Kubrick's films in the last 30 years have come when he has turned his actor's faces into  Think of Malcolm McDowell in A Clockwork Orange (1971), Jack Nicholson in The Shining (1980), and Vincent D'Onofrio in Full Metal Jacket (1987).\\
    
    \rule{0pt}{4ex}& \multirow{3}{*}{\vspace{.65cm}Vu} 
    & \textsc{sl.} & Critics laud Merchant-Ivory's exit from the 19th century in this adaptation of a semiautobiographical novel by Kaylie Jones (daughter of novelist James Jones).  \\
    \rule{0pt}{2.5ex}& & \textsc{sl.} & Mercer was the lifelong love of Franklin Roosevelt, and the revelation of their affair nearly ended his marriage to Eleanor. \\
    \rule{0pt}{2.5ex}& & \textsc{sl.} & Critics find the book entertaining, praising digressions on gambling, laughing, and love, as well as Pinker's pop-culture references. \\
    
    \rule{0pt}{4.5ex}& \multirow{3}{*}{\vspace{.65cm}Yang} 
    & \textsc{sl.} & Critics laud Merchant-Ivory's exit from the 19th century in this adaptation of a semiautobiographical novel by Kaylie Jones (daughter of novelist James Jones). \\
    \rule{0pt}{2.5ex}& & \textsc{tr.} & Visitors are encouraged to come during daylight hours, when the park is safer and better patrolled by employees. \\
    \rule{0pt}{2.5ex}& & \textsc{tr.} & All Ireland loves a horse, and County Kildare can claim to be at the heart of horse country.\\
    
  \bottomrule
   \end{tabular}
   \caption{Showing the three nearest neighbors for example sentences from a random 1,000-sample subset of the \textit{matched} test set. All results are for single (non-ensemble) models. The genres have been abbreviated.}
   \label{knnExample} 
\end{table*}

\begin{figure*}[t]
  \includegraphics[width=\textwidth]{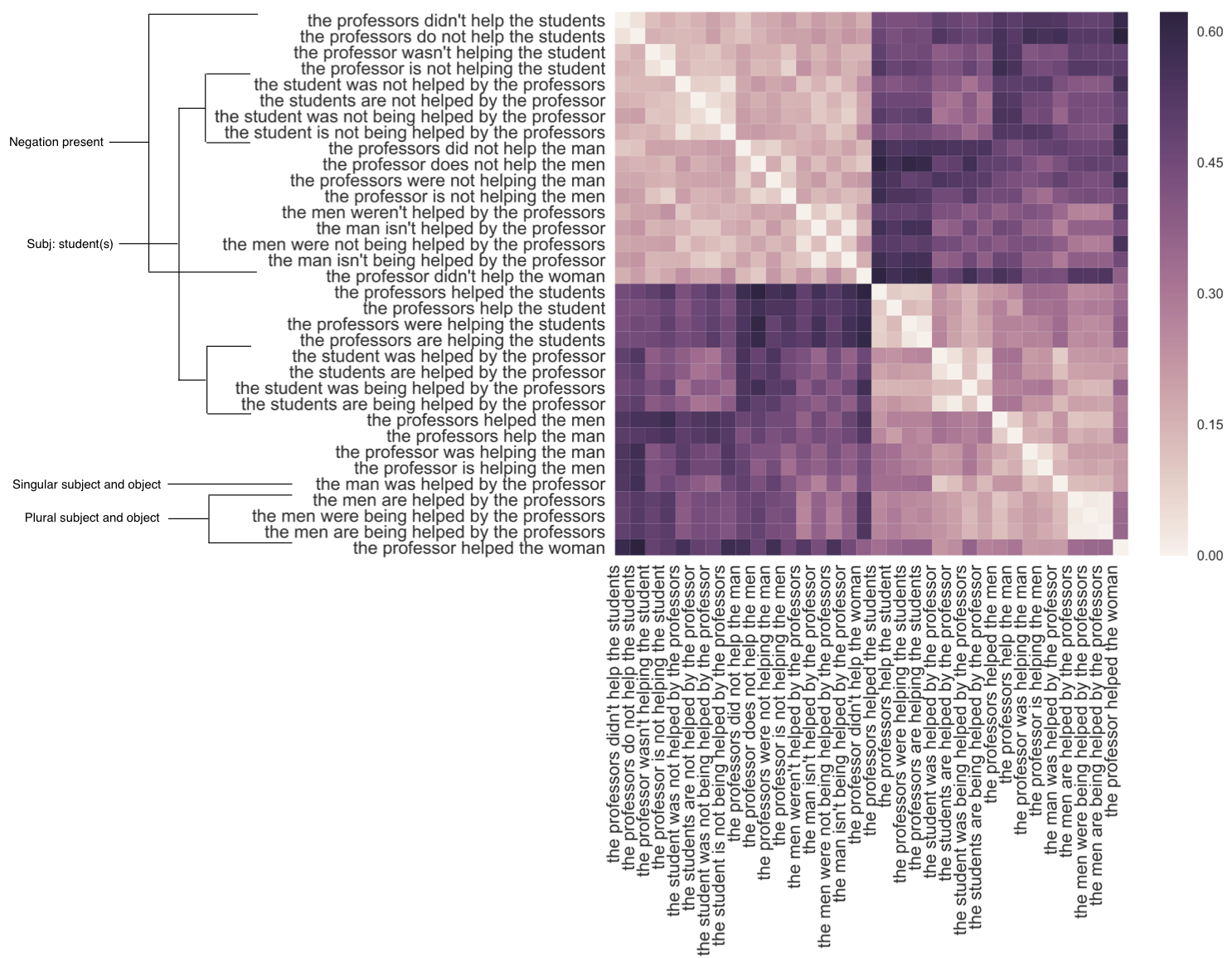}
  \caption{A heatmap showing the cosine similarity between sentence vectors. The vectors were rendered by the \citeauthor{nie2017stacked}~model. The plots for other systems are very similar.} 
  \label{cosine}
\end{figure*}

\subsection{Nearest Neighbors}

\paragraph{Test Set Sentences} The competition participants were asked to submit sentence vectors for all the premise and hypothesis sentences in the test sets. We randomly sample 1,000 sentences from the \textit{matched} test set and, using cosine similarity, calculate their pairwise distances against all sentences in the \textit{matched} test set. Table~\ref{1nn} shows the percentage of times the first nearest neighbor belongs to the same genre as the chosen sentence. All models score fairly highly on this metric, suggesting that the learned representations are not genre-agnostic, despite their effectiveness in unseen genres. The models with higher percentage accuracy on the NLI task (see Table~\ref{results}) show better performance on this metric as well, suggesting that this genre clustering property correlates with the overall quality of the metric space that each model uses to represent sentences.

The better models are also more interpretable. Table~\ref{knnExample} shows example sentences and their three nearest neighbors for all models. It appears that entity identity is important for the \citeauthor{nie2017stacked}~model, though not it a way that is tied to syntactic position. For the \textit{Critics loved Merchant-Ivory} example, we see matches to critics. In the \textit{Students love the rich culture} example, we similarly see many matches to school and love. Since for each premise sentence in the MultiNLI corpus, there are 3 associated hypothesis sentences, it's not surprising to see that the first nearest neighbor is often one of these associated sentences, like in the \textit{Critics} example where the first nearest neighbor for all systems is the premise sentence. We found that for some examples, the better performing systems like \citeauthor{nie2017stacked}'s~had all three associated sentences as their top three nearest neighbors.

\paragraph{Probe Sentences}

During the competition, we additionally provided a set of automatically generated probe sentences meant to aid error analysis. These probe sentences are produced to vary along dimensions relevant to probing for semantic role and negation information. We asked submitting teams to supply vectors for these sentences in addition to those in the test set. Figure~\ref{cosine} shows the cosine similarity between a subset of these sentence vectors rendered by \citeauthor{nie2017stacked}'s \citeyearpar{nie2017stacked} system.  We find that all systems (except that of \citeauthor{balazs2017refining}, who did not submit these vectors) show similar behavior on these sentences, and we do not observe a clear correlation between behavior here and model performance. Perhaps unsurprisingly, we observe that sentences tend to be more similar to one another the more structural features they have in common. We observe this clearly for negation, identity of the subject, and tense, though continuous tenses are not reliably differentiated from others.

\section{Conclusion}
We find that BiLSTM-based models with max pooling or intra-sentence attention represent a popular and effective strategy for sentence encoding, and that systems based on this technique perform very well at the task of NLI. 

We note that all submitted systems performed reasonably well across the many subsets of the data reflected by our supplementary tags, suggesting that none of these models exploit any particular narrow feature of the task or data to perform well. We also note that model performance does not vary much between the \textit{matched} and \textit{mismatched} sections of the test set. This means that submitted systems are likely capturing reasonably general strategies for extracting representations of meaning from text. As the systems get better, and fit the training data more closely, the disparity between \textit{matched} and \textit{mismatched} sets may appear. Both of these findings, though, bolster our expectation that the best of the submitted systems represent some of the best general-purpose architectures for sentence encoding available. 

However, the task of NLI is far from being solved, and no submitted system approaches human performance, suggesting that there is ample room for further research on both the task and on the more general problem of sentence representation learning. Since many of the examples in MultiNLI require substantial commonsense background knowledge to solve fully, we suspect that the use of large outside datasets and resources (labeled or otherwise) will be crucial to making substantial further progress in this setting. 

\section*{Acknowledgments}

This work was made possible by a Google Faculty Research Award to Sam Bowman and Angeliki Lazaridou, and was also supported by a gift from Tencent Holdings. Allyson Ettinger contributed the supplementary probe sentences. We also thank George Dahl and the organizers of the RepEval 2016 and RepEval 2017 workshops for their help and advice. 

\bibliography{MLSemantics}
\bibliographystyle{emnlp_natbib}

\end{document}